\documentclass[a4paper,conference]{IEEEtran}
\ifCLASSINFOpdf
   \usepackage[pdftex]{graphicx}
\else
\fi
\usepackage{amsmath}
\ifCLASSOPTIONcompsoc
 \usepackage[caption=false,font=normalsize,labelfont=sf,textfont=sf]{subfig}
\else
 \usepackage[caption=false,font=footnotesize]{subfig}
\fi
\usepackage{url}

\usepackage{multirow}
\usepackage{amssymb}

\def\chk{\checkmark}

\begin{document}
%
\title{Monocular Depth Estimation Using Cues Inspired by Biological Vision Systems}

\author{
    \IEEEauthorblockN{Dylan Auty, Krystian Mikolajczyk}
    \IEEEauthorblockA{
        MatchLab Group\\
        Electrical and Electronic Engineering Department\\
        Imperial College London\\
        Email: dylan.auty12@imperial.ac.uk, k.mikolajczyk@imperial.ac.uk
    }
}


\maketitle


%
\IEEEpeerreviewmaketitle

\begin{abstract}
    Monocular depth estimation (MDE) aims to transform an RGB image of a scene into a pixelwise depth map from the same camera view. It is fundamentally ill-posed due to missing information: any single image can have been taken from many possible 3D scenes.
    Part of the MDE task is, therefore, to learn which visual cues in the image can be used for depth estimation, and how. With training data limited by cost of annotation or network capacity limited by computational power, this is challenging.
    
    In this work we demonstrate that \textit{explicitly} injecting visual cue information into the model is beneficial for depth estimation. Following research into biological vision systems, we focus on semantic information and prior knowledge of object sizes and their relations, to emulate the biological cues of relative size, familiar size, and absolute size. We use state-of-the-art semantic and instance segmentation models to provide external information, and exploit language embeddings to encode relational information between classes. We also provide a prior on the average real-world size of objects. This external information overcomes the limitation in data availability, and ensures that the limited capacity of a given network is focused on known-helpful cues, therefore improving performance.
    We experimentally validate our hypothesis and evaluate the proposed model on the widely used NYUD2 indoor depth estimation benchmark. The results show improvements in depth prediction when the semantic information, size prior and instance size are explicitly provided along with the RGB images, and our method can be easily adapted to any depth estimation system.
\end{abstract}
\section{Introduction}
\label{sec:introduction}
    Monocular depth estimation (MDE) is the task of inferring, from a single still image, the distance from the camera to every point in the scene visible in the image. This is non-trivial, since there exist many possible 3D scenes from which any given image may have been taken. Despite this, many machine learning (ML) systems have been able to achieve good performance in this task. These methods leverage prior knowledge gleaned from data to produce plausible results, though this knowledge is often acquired in a nonspecific, implicit way from large training datasets.
    
    Similarly, biological vision systems also perform MDE well, despite being subjected to the same limitations as machine learning approaches. Several biological cues have been identified \cite{hershenson_pictorial_1998} that aid depth estimation in humans and some animals; removing or perturbing these cues harms performance \cite{wagner_barn_1991, nagata_depth_2012, harkness_chameleons_1977, sousa_judging_2011}. These cues are, as in ML systems, learned by exposure to many examples. 
    In an ML context however, a lack of suitable multi-task-labelled data makes the learning of these cues hard.
    
    To overcome this limitation, we present a method of explicitly introducing information that emulates three known biological depth cues, without the need for multi-task-labelled data. This provides prior information that is known to aid MDE performance, that would otherwise be difficult to learn implicitly from a limited dataset with limited model capacity. 
    To the best of our knowledge, none of the modern MDE methods explicitly incorporate external data using language models, with a view to replicating biological depth cues.
   
    Our main contributions are:
    \begin{itemize}
        \item We present a novel approach to exploit semantic and instance information that is known to be used in biological depth cues to improve performance of machine-learning MDE systems,
        \item We encode this information via natural language models, to include prior knowledge about the nature of a given object and its relation to others in the world,
        \item Our method improves performance significantly over the baseline, overcoming limitations in both data availability and model capacity by injecting a prior to focus learning to the most important image features.
    \end{itemize}
\section{Related Work}
\label{sec:relatedwork}
    \textbf{Biological Depth Perception.} In biological vision, research has identified several explicit monocular depth cues. The apparent sizes of objects relative to one another and to their known actual dimensions provide strong cues \cite{hershenson_pictorial_1998, sousa_judging_2011}, as do texture gradient and perspective. Some less intuitive biological cues include defocus \cite{nagata_depth_2012}, accommodation \cite{wagner_barn_1991, harkness_chameleons_1977}, and the scattering of light through the atmosphere over long distances \cite{cozman_depth_1997}.
    In interpreting these cues for use in ML, we note that biological and ML-based vision systems are not known for certain to work in the same way. However, there is nonetheless some evidence that biological depth cues may be being exploited by some deep learning systems.
    
    \noindent\textbf{Monocular Depth Estimation.} Deep learning methods have achieved good performance in MDE. They often consist of an encoder backbone such as ResNet \cite{he_deep_2016}, AlexNet \cite{krizhevsky_imagenet_2012}, or VGG \cite{simonyan_very_2015}, and a decoder that expands the extracted features to the desired dimensions of the depth map. The problem, however, is ill-posed: there exist many possible 3D scenes that could have produced the provided input image. 
    Therefore, any MDE system must learn from training data to spot appropriate visual depth cues from wider image context, and to understand what predictions are reasonable. We posit that the cues and prior knowledge learned by such systems will be the same as in the biological domain.
    
    Many methods have attempted to include wider context. \cite{eigen_depth_2014, eigen_predicting_2015, lee_big_2019} use a multi-scale network that predicts coarse and then progressively finer detail in the final depth map. \cite{wang_pixel2mesh_2018} apply progressive refinement to a mesh deformation task, and \cite{chen_deeplab:_2017,chen_attention-based_2019} used stacked atrous convolutions to gather context for use in semantic segmentation.
    Other methods to learn cues from context instead co-predict depth with other tasks, such as semantic labels, occluding contours, or surface normals \cite{ramamonjisoa_sharpnet:_2019, jiao_look_2018, bai_monocular_2019}. By sharing weights or aligning predictions, the model is explicitly encouraged to pay attention to image features that are relevant to all tasks, which it may not have done otherwise. 
    Semantic/depth co-prediction is also used in multi-view MDE \cite{chen_towards_2019, jung_fine-grained_2021} to positive effect, though as these methods are self-supervised and multi-view, they cannot be directly compared to supervised, single-view MDE. All such co-prediction methods require depth \& semantic labels for data, and must by design expend model capacity on an auxiliary task that is not the end goal of MDE.
    
    Alternatively, information can be introduced from external sources. This can be especially useful if training data is labelled for only a single task, making multi-task training infeasible. This information can come from the dataset or from other models, such as the semantic labels used in \cite{li_learning_2021, lopez-rodriguez_desc_2020, jung_fine-grained_2021}. External information can also come in the form of useful assumptions about the world. \cite{van_dijk_how_2019} found that the good performance of MonoDepth \cite{godard_unsupervised_2017} was partly because the height in an image of a given car was a much stronger indicator of depth than the apparent size of that car. \cite{lee_big_2019} explicitly emphasise an assumption that the world is locally comprised of planar surfaces. \cite{casser_depth_2018, lopez-rodriguez_desc_2020} exploit the known relationship between apparent height, camera intrinsics, and true height of an object, with both methods allowing the true height to be learned by the model.
    \cite{wang_sdc-depth_2020} note that depths are similar for pixels sharing the same semantic label, and leverage this to divide the MDE task among multiple class-specific MDE models instead of a single general-purpose one.
\section{Method}
\label{sec:method}
        \begin{figure}[!t]
            \centering
            \subfloat[RGB]{\includegraphics[width=0.12\textwidth]{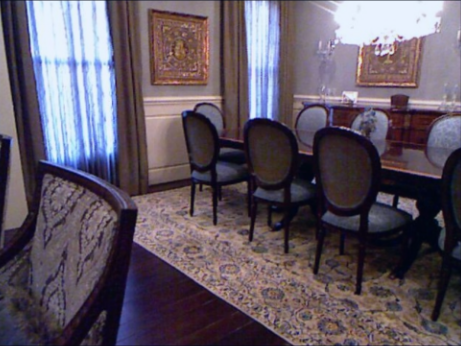}%
            \label{fig:depthCues_1_rgb}}
            \hfil
            \subfloat[Instance map]{\includegraphics[width=0.12\textwidth]{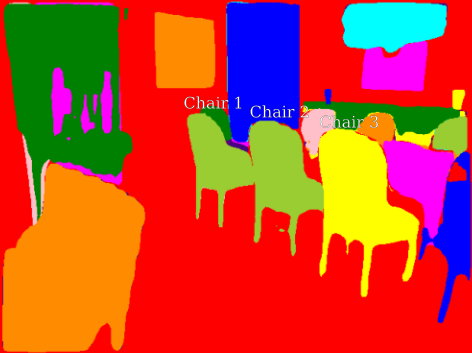}%
            \label{fig:depthCues_1_sem}}
            \hfil
            \subfloat[RGB]{\includegraphics[width=0.12\textwidth]{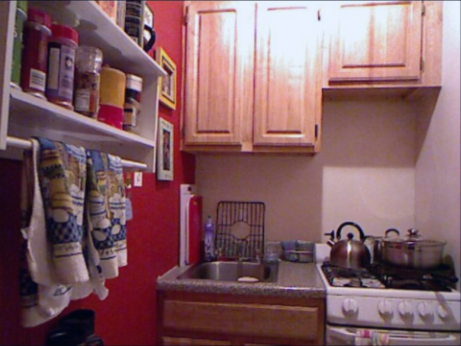}%
            \label{fig:depthCues_2_rgb}}
            \hfil
            \subfloat[Instance map]{\includegraphics[width=0.12\textwidth]{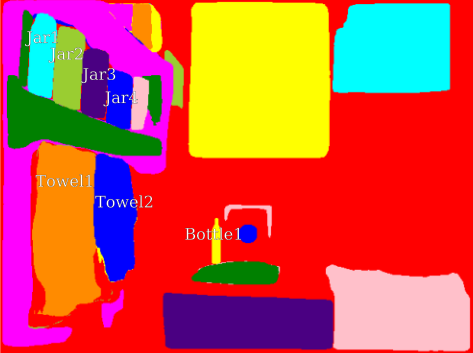}%
            \label{fig:depthCues_2_sem}}
            \caption{Images illustrating size-related biological depth cues. In fig. \ref{fig:depthCues_1_rgb}/\ref{fig:depthCues_1_sem}, relative size is clear in the three identical chairs: the chair on the left appears ``smaller" and also the furthest away, due to perspective. In fig. \ref{fig:depthCues_2_rgb}/\ref{fig:depthCues_2_sem}, familiar and absolute size can be seen: the bottle appears to be the same size as the jars on the left, but prior knowledge that the bottle is larger than a jar tells us that the bottle is further away. The towels, of unknown true size, appear close simply because they appear large. Please zoom in to see labels.}
        \label{fig:depthCues}
    \end{figure}
        
    \begin{figure*}[!t]
        \centering
        \includegraphics[width=0.8\textwidth]{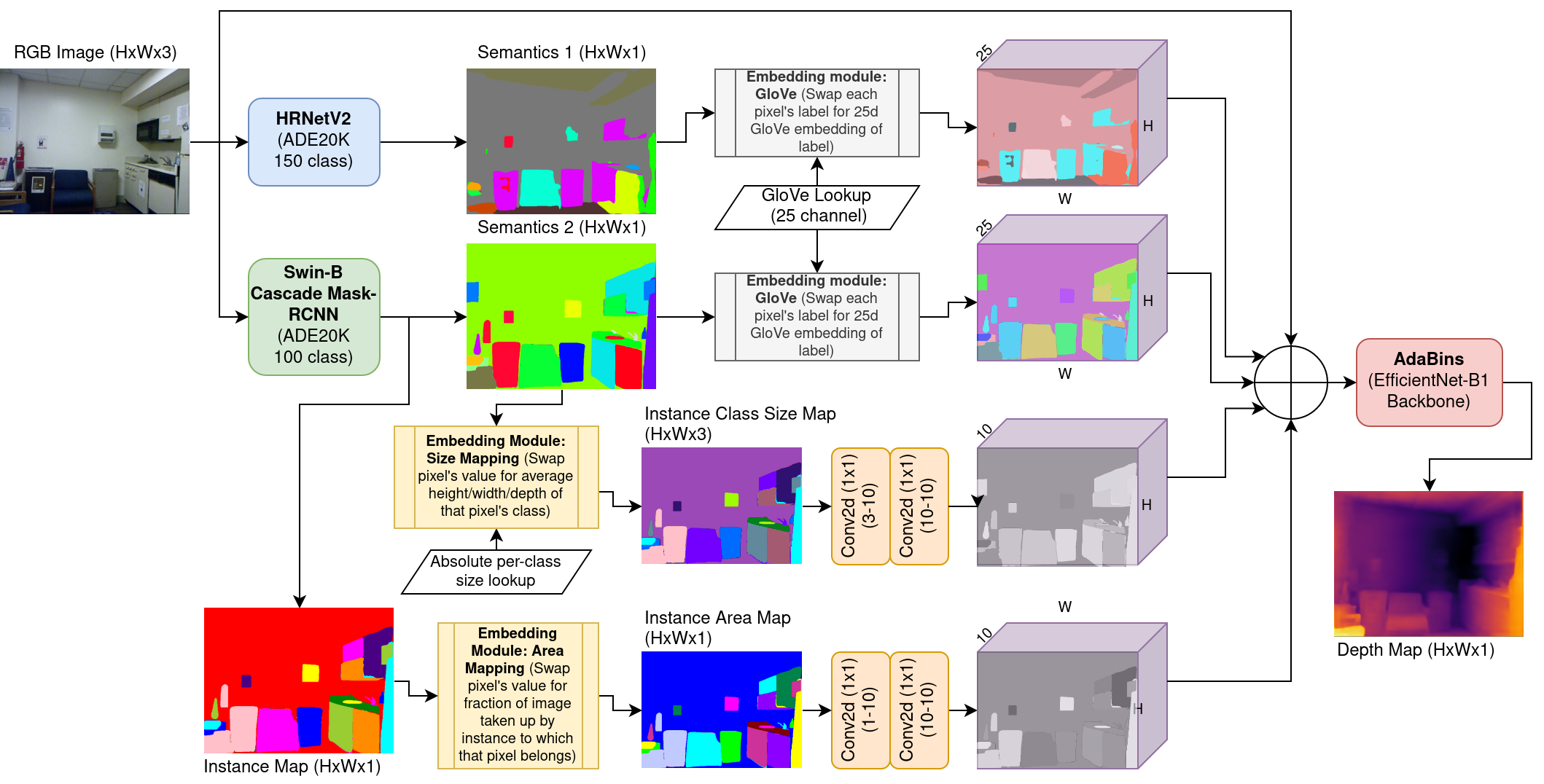}
        \caption{The proposed approach. HRNetV2 and Swin-B Cascade Mask-RCNN provide semantic and semantic + instance segmentation information respectively. Inference is run by each of these pretrained pipelines on each image from the depth dataset. Results are embedded, concatenated to the input image, and fed to the AdaBins with EfficientNet-B1 backbone to produce a depth map. Only the MDE model is permitted to learn at training time; all other models are fixed.}
        \label{fig:architecture}
    \end{figure*}

    Our method links three known biological depth cues with the concept of explicitly adding external information, specifically the cues of \textbf{relative size}, \textbf{familiar size}, and \textbf{absolute size}. These cues are known to aid performance in biological vision systems, and by introducing information that is known to be helpful for MDE, we ensure that the limited capacity of the network is spent on learning to infer depth from depth cues, and not on learning the depth cues themselves. We note the work of \cite{van_dijk_how_2019}, but we instead posit that a height prior will be less useful than a size prior in the indoor domain, due to the highly variable camera angles and scene geometries when compared to a driving dataset. Similar to both \cite{lopez-rodriguez_desc_2020} and \cite{casser_depth_2018}, we also incorporate average actual size of objects to facilitate familiar size cues, although our method is provided with the approximate actual dimensions of objects instead of needing to learn them. To account for intra-class variation, the model is permitted to learn to embed these dimensions. We also explicitly include instance size information, thereby facilitating both relative and absolute size cues. 
    
    \subsection{Explicit Depth Cue Information}
    \label{sec:explicit-depth-cue-information}

        
        Our method introduces semantic segmentation information, average real-world dimensions for each semantic label, instance segmentation information, and the size of a given instance as a proportion of the whole image (referred to here as "apparent size"). These pieces of information directly relate to the three known biological depth cues of \textbf{relative size}, \textbf{familiar size}, and \textbf{absolute size} \cite{hershenson_pictorial_1998, sousa_judging_2011}. These all rely in some way on the size of the object in the visual field, since a consequence of perspective is that objects will appear larger when closer to the viewer:
        
            \noindent\textbf{Relative size}: When two objects of the same semantic class are identified by the viewer, their apparent size (the proportion of the visual field that each instance takes up) may be compared. If one of them appears to be larger in the visual field than the other, despite a learned prior that these objects share the same absolute size, then this shows that one is closer to the camera than the other. 
            
            \noindent\textbf{Familiar size}: The apparent size of objects in the visual field is related to prior knowledge of the actual size of these objects (e.g. knowing how big an object ``should" be). By comparing the apparent and known actual sizes, the distance of the object from the viewer can be inferred.
            
            \noindent\textbf{Absolute size}: Less intuitively, one study has shown that even when the absolute size of an object is unknown to the viewer (e.g. an unknown object), the apparent size of that object still influences the viewer's estimation of the distance to that object \cite{sousa_judging_2011}. The authors argue that this is because there exists an underlying assumption of what absolute object sizes are generally more likely, despite the object's unfamiliarity.
        
        While a given deep-learning MDE system \textit{may} be able to infer some of this information from the training set, we hypothesise that, given that training data is often limited, it would be more efficient to introduce the information required to emulate these biological cues directly. Explicitly emphasising information that is known to increase depth performance in biological systems will, we expect, increase performance in machine learning systems.
    
    \subsection{Approach}
        
        
        Our proposed approach is shown in figure \ref{fig:architecture}.
        It is comprised of a semantic segmentation module, an instance segmentation module, an embedding module, and a depth estimation module.
        
        \subsubsection{Semantic and Instance Segmentation}
        \label{sec:semantic_and_instance_segmentation}
            Semantic labels and actual per-class dimensions are needed for the familiar size cue, and instance size information permits use of absolute and relative size cues.
            We obtain this information using off-the-shelf models for semantic and instance segmentation. To use inference from these models on NYUD2 as part of MDE is to incorporate the outside knowledge contained in them, since these models were not trained on NYUD2. In this way, we overcome the data limitation that prevents full exploitation of these depth cues, \textbf{without} having to annotate more data. A further effect of this is that the auxiliary models do not need to be retrained for each new target depth dataset, provided that dataset is in the same domain (e.g. indoor).
            
            Semantic information directly supports the familiar size cue, where the model relates the apparent size of objects to prior knowledge about their actual size. When combined with our embeddings, this provides external knowledge about the relationships between object classes (e.g. bigger than, smaller than, large, small), and may also provide a weak prior on the true dimensions of these objects. Instance segmentation provides more semantic information, but also provides explicit areas. Providing the areas of objects encourages use of the absolute size cue, where larger objects are interpreted as being closer to the viewer. Combining both instance area information and semantic labels for a given instance will permit the use of the relative size cue, where apparent sizes of objects known to be of the same class (or objects of different classes that, through embeddings, have been determined to be of similar actual dimensions) can be compared.
            
            The semantic segmentation model used was HRNetV2 \cite{sun_high-resolution_2019}, pretrained on the ADE20K dataset's \cite{zhou_semantic_2018} 150-class Scene Parsing Challenge subset \cite{zhou_scene_2017}. We denote the use of information from this model with $H_{As}$. Checkpoints were provided by the authors of \cite{zhou_semantic_2018}\footnote{Available at \url{http://sceneparsing.csail.mit.edu/model/pytorch/ade20k-hrnetv2-c1/}}. On the testing split of the same dataset, this model achieves a mean IoU of 43.20, and an 81.47\% pixel accuracy. 
            The instance segmentation model used was Cascade Mask R-CNN with a Swin-B backbone \cite{liu_swin_2021}, trained on the ADE20K dataset's 100-class Places Challenge instance segmentation subset. We denote the use of information from this model with $S_{Ai}$. Training was performed for 36 epochs on a single NVIDIA Titan RTX graphics card. The model achieved an instance bounding box mAP of 0.364, and instance segmentation mask mAP of 0.295. In some experiments, we also use Mask-RCNN trained on MSCOCO \cite{lin_microsoft_2015}, and we denote the use of information from this with $M_C$.
        
        \subsubsection{Embedding Module}
        \label{sec:embedding_module}
            The embedding module transforms the incoming raw information from external sources into a format that is more useful to the depth estimation module.
            
            In the case of the semantic labels, these are introduced in order to facilitate the \textbf{familiar size} cue. To do this, some prior knowledge of what an object actually is (i.e. what the semantic labels actually ``mean") is necessary, and is clearly more useful than an arbitrary class number alone. Therefore, GloVe language embeddings \cite{pennington_glove_2014} were used to represent class labels, specifically the 25-dimensional pre-trained word vectors trained on a corpus of 2 billion tweets. These embeddings may not introduce information about actual object dimensions in the world, but may still encode relational actual size information to other classes. In rare cases, very weak priors for actual size \textit{may} be present. We verify the efficacy of the GloVe embeddings in section \ref{sec:ablation}.
            
            To improve on the weak actual size prior encoded by GloVe, we also input embedded approximate spatial dimensions in metres, intended to be representative of the average size of objects of that class. These were manually estimated and thus cannot be totally accurate, but we allow the network to learn a 10-dimensional embedding of this absolute size that will, over training, come to encompass information about the distribution of actual object sizes across different instances. This is similar to \cite{lopez-rodriguez_desc_2020} and \cite{casser_depth_2018}, but provides a starting point rather than forcing the network to learn from nothing. We denote use of these sizes (in the form of a pixelwise map) by $Sz$ or $Size$.
            
            In the case of instance sizes, we first normalise the area in pixels of a given instance to represent the fraction of the image it takes up, then make use of two 1x1 convolution layers to allow the network to transform the scalar instance size to a learned 10-dimensional size embedding. The use of instance size information is denoted by either $S_{Ai}(M)$ or $S_{Ai}(B)$, where $M$ and $B$ denote the use of the pixelwise mask or the bounding box respectively when calculating the area of an instance. The final output is a pixelwise map of embeddings, where each pixel's embeddings correspond to information about that pixel (i.e. embedding of size of object class at that pixel, embedding of apparent size of the instance to which that pixel belongs, GloVe embedding of that pixel's semantic class). This final output is concatenated to the input image, and the result is fed to the depth estimation model.
        
        \subsubsection{Depth Estimation Module}
        \label{sec:depth-estimation-module}
            The Depth Estimation Module follows the architecture and setup of AdaBins \cite{bhat_adabins_2020} very closely. This method comprises an EfficientNet-B5 \cite{tan_efficientnet_2019} encoder/decoder, whose final pooling and classification layers are stripped off so that it outputs 128-dimensional image features instead of a depth map. These dense features are then fed to the AdaBins module itself, which performs ordinal regression based on these features into 256 bins whose centres and widths it learns to modify based on the input image. Finally, to produce a smooth depth map, pixel depth values are given by an interpolation between the centres of the most likely bin assignments for that pixel.
            
            In our pipeline, GPU RAM constraints forced us to replace the EfficientNet-B5 encoder/decoder with an EfficientNet-B1 to permit training; this makes the network considerably smaller at the cost of a decrease in performance. Training with the EfficientNet-B5 backbone and AdaBins module together was not physically possible.
            EfficientNet-B5 achieves 83.6\% top-1 accuracy on ImageNet classification with 30M parameters, whereas EfficientNet-B1 achieves 79.1\% top-1 accuracy on the same task with only 7.8M parameters. 
            The difference in performance this causes can be seen in table \ref{tab:results1} by comparing the AdaBins-B5 results to our baseline. Because of this forced reduction in network capacity, we compare to our own baseline in our experiments, and provide SOTA results only for context.
            
            The input image is concatenated with features from the embedding module, before being fed to the depth estimation module.
            Accordingly, when training the Depth Estimation Module, we begin with a 3-channel input EfficientNet-B1 pretrained on ImageNet, whose first layer we extend to have as many channels as required for a given experiment. These new channel weights are randomly initialised.
\section{Experiments}
\label{sec:experiments}
    \begin{table*}[t!]
        \footnotesize
        \centering
        \caption{Performance of our best models against the baseline, with some results from SOTA models for context. Significant improvement over baseline can be seen when using all available auxiliary information. Notation used is described in sections \ref{sec:semantic_and_instance_segmentation} and \ref{sec:embedding_module}. Best results in bold, second-best underlined. Note that our methods use a much smaller backbone than AdaBins-B5, due to GPU RAM constraints.}
        \begin{tabular}{l|llll|lll}
            Method & \multicolumn{4}{l|}{Distance metrics ($\downarrow$)} & \multicolumn{3}{l}{$\delta$ accuracy ($\uparrow$)} \\
                                                  & Abs & Sq   & RMS   & RMSL   & $\delta_1$  & $\delta_2$ & $\delta_3$ \\ \hline
            \hline
            
            Baseline (AdaBins-B1)           & 0.123     & 0.076    & 0.413 & 0.151  & 0.864   & 0.976    & 0.995 \\
            Ours ($S_{Ai}+H_{As}+S_{Ai}(B)+Size$)
                                                & 0.115    & 0.065    & \underline{0.385} & \underline{0.140} & \underline{0.888}   & \underline{0.982}    & \underline{0.996} \\ 
            Ours ($S_{Ai}+H_{As}+S_{Ai}(M)+Size$)
                                                & 0.114     & \underline{0.064}    & 0.386 & \underline{0.140}  & 0.885   & \underline{0.982}    & \underline{0.996} \\ 
            \hline
            DORN \cite{fu_deep_2018}              & 0.115     & -        & 0.509 & -   & 0.828   & 0.965    & 0.992 \\
            BTS \cite{lee_big_2019}              & \underline{0.110}     & -        & 0.392 &  -     & 0.885   & 0.978    & 0.994 \\
            SDC-Depth \cite{wang_sdc-depth_2020} & 0.128 & - & 0.497 & 0.174 & 0.845 & 0.966 & 0.990 \\
            AdaBins-B5 \cite{bhat_adabins_2020}  & \textbf{0.104}     & \textbf{0.057}    & \textbf{0.365} & \textbf{0.131}  & \textbf{0.902}   & \textbf{0.983}    & \textbf{0.997} \\\hline
        \end{tabular}
        \vspace{1em}

        \label{tab:results1}
    \end{table*}
    
    \begin{table*}[h!]
        \centering
        \caption{Comparing performance when including different auxiliary information in input to AdaBins-B1 model. Best results for each metric highlighted in bold. Notation is described in sections \ref{sec:semantic_and_instance_segmentation} and \ref{sec:embedding_module}.}
        \begin{tabular}{llll|llll|lll}
        \multicolumn{4}{l|}{Method} & \multicolumn{4}{l|}{Distance metrics ($\downarrow$)} & \multicolumn{3}{l}{$\delta$ accuracy ($\uparrow$)} \\
         $Sem_1$   & $Sem_2$   & $Area$     & $Size$ & Abs Rel   & Sq Rel   & RMS   & RMSL   & $\delta_1$  & $\delta_2$ & $\delta_3$ \\ \hline\hline
        \multicolumn{4}{c|}{Baseline (AdaBins-B1)}
                                              & 0.123     & 0.076    & 0.413 & 0.151  & 0.864   & 0.976    & 0.995 \\
        $H_{As}$ &          &          &      & 0.115     & 0.067    & 0.394 & 0.143  & 0.884   & 0.981    & 0.995 \\    
        $S_{Ai}$ &          &          &      & 0.119     & 0.068    & 0.394 & 0.145  & 0.877   & 0.980    & \textbf{0.996} \\    
        $S_{Ai}$ &          & $S_{Ai}(M)$ &      & 0.119     & 0.070    & 0.397 & 0.146  & 0.875   & 0.980    & 0.995 \\    
        $S_{Ai}$ &          & $S_{Ai}(B)$ &     & 0.121     & 0.070    & 0.399 & 0.147  & 0.872   & 0.979    & 0.995 \\ 
        $M_C$    &          & $M_C$(M)    &      & 0.121     & 0.073    & 0.405 & 0.148  & 0.872   & 0.978    & 0.995 \\    
        $M_C$    & $H_{As}$ & $M_C$(M)    &      & 0.117     & 0.070    & 0.396 & 0.144  & 0.878   & 0.980    & \textbf{0.996} \\
        $S_{Ai}$ & $H_{As}$ & $S_{Ai}(M)$ &      & 0.115     & 0.066    & 0.386 & 0.141  & 0.884   & \textbf{0.982}    & \textbf{0.996} \\ 
        {$S_{Ai}$} &          &          & {\chk} & {0.120} & {0.069} & {0.397} & {0.146} & {0.874} & {0.981} & {0.996} \\ 
        {$S_{Ai}$} &          & {$S_{Ai}(M)$} & {\chk} & {0.119}     & {0.069}    & {0.396} & {0.146}  & {0.875}   & {0.980}    & {\textbf{0.996}} \\ 
        $S_{Ai}$ &          & $S_{Ai}(B)$ & \chk & 0.120    & 0.068    & 0.397 & 0.146  & 0.871   & 0.980    & \textbf{0.996}\\ 
        $S_{Ai}$ & $H_{As}$ & $S_{Ai}(M)$ & \chk & \textbf{0.114 }    & \textbf{0.064}    & 0.386 & \textbf{0.140}  & 0.885   & \textbf{0.982}    & \textbf{0.996} \\ 
        $S_{Ai}$ & $H_{As}$ & $S_{Ai}(B)$ & \chk & 0.115    & 0.065    & \textbf{0.385} & \textbf{0.140}  & \textbf{0.888}   & \textbf{0.982}    & \textbf{0.996} \\ 
        \hline
        \multicolumn{4}{c|}{Baseline (no semantics)}               & {0.123}     & {0.076}    & {0.413} & {0.151}  & {0.864}   & {0.976}    & {0.995} \\
        
        \multicolumn{4}{c|}{{Raw semantic labels}}                   & \textbf{0.119}     & 0.070    & 0.398 & 0.146  & 0.875   & \textbf{0.980}    & \textbf{0.996} \\ 
        
        \multicolumn{4}{c|}{{Random 25-d labels}}                    & {\textbf{0.119}}     & {0.070}    & {0.395} & {\textbf{0.145}}  & {\textbf{0.878}}   & {0.979}    & {0.995} \\ 
        \multicolumn{4}{c|}{{25-d GloVe embeddings}}                 & {\textbf{0.119}}     & {\textbf{0.068}}    & {\textbf{0.394}} & {\textbf{0.145}}  & {0.877}   & {\textbf{0.980}}    & {\textbf{0.996}} \\ 
        \hline  
        \end{tabular}
        \vspace{1em}

        \label{tab:results2}
    \end{table*}
    
    In this section we first present the experimental setup, then discuss the overall performance and the contributions of individual components.
    \subsection{Experimental Setup}
    \label{sec:experimental_setup}
    All of our experiments were on the NYU Depth Dataset V2 (NYUD2) \cite{silberman_indoor_2012}. As it appears that slightly different versions of this dataset have emerged over time, we note that the copy used was that provided with the official codebase of \cite{lee_big_2019}\footnote{See \url{https://github.com/cogaplex-bts/bts/tree/master/pytorch}}. Our code\footnote{\url{https://github.com/DylanAuty/MDE-biological-vision-systems}} is a fork of the official implementation of \cite{bhat_adabins_2020}, and is written using PyTorch.
    {Due to GPU RAM constraints, we were forced to change the backbone from EfficientNet-B5 to EfficientNet-B1; this is explained further in section \ref{sec:depth-estimation-module}.}
    All depth-estimation experiments were trained with batch size of 9 on a system with two NVIDIA GeForce GTX 1080 graphics cards {(8GB memory each)}. The semantic and instance segmentation models and checkpoints are described in detail in section \ref{sec:semantic_and_instance_segmentation}.
    
    The loss function used is exactly that used in AdaBins, and consists of a pixel-wise scale-invariant depth loss and a bin-center density loss component. The former enforces that depth predictions are accurate, and the latter that the adaptive bins chosen for a given image will have centres whose distribution matches that of the input image. Full details can be found in \cite{bhat_adabins_2020}.
    
    For metrics, we follow common convention in the field of using the six metrics defined by \cite{eigen_depth_2014}: Abs relative difference (Abs or Abs Rel): $\frac{1}{T}\sum_{i=1}^{T} \frac{|y_i - y_i^*|}{y_i^*}$, Squared relative difference (Sq or Sq Rel): $\frac{1}{T}\sum_{i=1}^{T} \frac{\|y_i - y_i^*\|}{y_i^*}$, lin. RMSE (RMS): $\sqrt{\frac{1}{T}\sum_{i=0}^{T}\|y_i - y_i^*\|^2}$, log RMSE (RMSL): $\sqrt{\frac{1}{T}\sum_{i=0}^{T}\|log(y_i) - log(y_i^*)\|^2}$, and the threshold accuracy $\delta_n$: \(\%\) of \(y_i\) s.t. $max(\frac{y_i}{y_i^*}, \frac{y_i^*}{y_i}) = \delta < thr$, where $\delta_n$ denotes that $thr = 1.25^n$ (we use $n \in \{1, 2, 3\}$).
    
    \subsection{Results}
        
        
        \begin{table*}[!t]
            \footnotesize
            \centering
            \caption{Ablation with different architecture: EfficientNet-B5 encoder + standard upsampling decoder w/ single-channel (depth) output. The AdaBins module is not present in this model. Notation in table is the same as in table \ref{tab:results2}.}
            \begin{tabular}{lll|llll|lll}
            \multicolumn{3}{l|}{Method} & \multicolumn{4}{l|}{Distance metrics ($\downarrow$)} & \multicolumn{3}{l}{$\delta$ accuracy ($\uparrow$)} \\
             $Sem_1$  & $Area$     & $Size$ & Abs Rel   & Sq Rel   & RMS   & RMSL   & $\delta_1$  & $\delta_2$ & $\delta_3$ \\ \hline\hline
            \multicolumn{3}{c|}{Alt. Baseline (EfficientNet-B5)}
                                        & 0.106 & 0.059 & 0.373 & 0.135 & 0.892 & \textbf{0.983} & \textbf{0.996} \\ 
    
            $S_{Ai}$ &          &         & \textbf{0.105} & 0.060 & 0.370 & \textbf{0.134} & 0.894 & 0.982 & \textbf{0.996} \\ 
            
            $S_{Ai}$ & $S_{Ai}(M)$ &      & 0.107 & 0.060 & 0.370 & 0.135 & 0.894 & \textbf{0.983} & \textbf{0.996} \\ 
            
            $S_{Ai}$ & $S_{Ai}(M)$ & \chk & 0.107 & \textbf{0.058} & \textbf{0.368} & \textbf{0.134} & \textbf{0.895} & \textbf{0.983} & \textbf{0.996} \\ 

            \hline  
            \end{tabular}
            \vspace{1em}
            \label{tab:resultsNewBaseline}
        \end{table*}
        
        Our primary results are presented in table \ref{tab:results1}, where we show that our best model achieves a significant improvement over our baseline. We also present three SOTA methods for additional context, though we emphasise that our method is not expected to compete with these directly due to the considerably smaller backbone that was used due to GPU RAM constraints (see section \ref{sec:depth-estimation-module}). Our baseline is identical to the AdaBins method but uses the smaller EfficientNet-B1 encoder, with an accompanying tradeoff in performance. The current SOTA (AdaBins, \cite{bhat_adabins_2020}) uses the much larger EfficientNet-B5 encoder that we were physically unable to train along with the AdaBins module, and that performs better than our baseline method as a result.
        
        It can be clearly seen that our method improves significantly on the baseline method, and outperforms \cite{fu_deep_2018} and \cite{wang_sdc-depth_2020}. Our method's semantic/instance segmentation models are frozen, so it does not require depth/semantic label pairs as \cite{wang_sdc-depth_2020} does.
        Compared to BTS \cite{lee_big_2019}, we use a significantly smaller backbone network. BTS also incorporate an assumption of local planarity and a multi-scale approach, both of which we do not use. We expect that if the GPU RAM limitations faced in this work were removed, further improvements over the SOTA method would likely be seen by applying the same method to the larger model of \cite{bhat_adabins_2020}. Our method is adaptable and may be applied to any MDE model, and we strongly believe that performance would improve where these models suffer from a lack of data or a lack of capacity to implicitly learn depth cues.
        
    \subsection{Ablation study}
        \label{sec:ablation}
        
        In designing the pipeline presented in section \ref{sec:method}, experiments were performed that varied the source of the semantic and instance segmentation and the encoding strategies used for this information, and that verified the efficacy of each of these pieces of information through the lens of the biological depth cues of familiar size, relative size, and absolute size. In table \ref{tab:results2} we show these different experiments in detail, with our baseline method included for comparison. Firstly, and perhaps most significantly, it can be clearly seen that no matter the type or source of extra information, the inclusion of \textit{any} additional explicit information to the network helps performance, showing strong evidence of deficiency in data or capacity without our method.
        
        \noindent\textbf{Source of semantics.} Semantic segmentation information was provided by different models, trained on different datasets. These were: HRNetV2 \cite{sun_high-resolution_2019} trained on the 150-class ADE20K semantic segmentation subset (termed $H_{As}$ in the results), Cascade Mask-RCNN with Swin-B backbone \cite{liu_swin_2021} trained as an instance segmentation model on the 100-class ADE20K "Places Challenge" instance segmentation subset (termed $S_{Ai}$ in results), and Mask-RCNN \cite{he_mask_2018} trained for semantic segmentation on MS-COCO \cite{lin_microsoft_2015}. In all cases, class label names were provided to the network as GloVe embeddings, as detailed in section \ref{sec:method}. We show that semantic information is better than no information at all, and that semantic information from HRNetV2 improves performance better than other sources tried; this may be due to better detections, or from the greater number of classes in the training dataset for this model. Further to this, our results clearly show that multiple sources of semantics perform better than just one; when including information from multiple semantic segmentation models, with all other factors remaining the same, performance was always improved.
        
        \noindent\textbf{Instance area and average object dimensions.} Instance area is included to explicitly emulate the absolute and relative size cues, and to aid familiar size.
        We find that using bounding box area instead of mask area produces worse performance, but that using instance area does not improve performance except as part of our best model. Our best model does therefore uses instance mask area.
        
        Actual per-class dimensions in metres are provided to the network that are representative of the average real-world size of a given object class, therefore facilitating the biological cue of familiar size. When introducing actual dimensions, performance improved in all cases. This is despite these dimensions not capturing the variance of object sizes within a given class; the implication of this is that the network was able to learn this variance on its own from the training data.

        \noindent\textbf{Alternative Baseline.}
        To confirm the findings discussed above, we perform similar experiments with a different baseline that consists of an encoder/decoder architecture, using an EfficientNet-B5 encoder \cite{tan_efficientnet_2019} and a standard upsampling decoder. The decoder outputs a single channel, which is passed through a ReLU with an offset of 0.0001 to permit use of the SILog loss function. The bin loss is removed. The results, in table \ref{tab:resultsNewBaseline}, show that the \textbf{joint} addition of the instance areas and per-class true sizes improve performance somewhat, which provides some evidence for the significance of the familiar size cue in particular.
        
        \noindent\textbf{Efficacy of GloVe embeddings}.
        The use of language embeddings is intended to convey more information than just numerical labels alone, and while the intention is that some small size-related information pertaining to relative and familiar size may be present, it is not immediately intuitive that it would be. To prove that the GloVe embeddings aid performance, therefore, an ablation was performed using different representations of the same semantic data. These results are presented at the bottom of table \ref{tab:results2}.
        
        Raw semantic labels are 1-channel integers representing the class number output by the segmentation network. The random embeddings used are 25-channel, and each value is drawn from a normal distribution whose mean and standard deviation matches that of the set of GloVe embeddings corresponding to the semantic classes used. These random embeddings remain constant throughout training (i.e. they are generated and assigned to a class, and then stored for future use).
        
        It may be seen in table \ref{tab:results2} that any embedding used performs better than either the raw semantics and the baseline with no semantics at all. Further to this, we demonstrate that language embeddings do perform marginally better than the random embeddings.

\section{Conclusion and Future Work}
\label{sec:conclusion}

    In this paper, we have shown that the simple introduction of semantic segmentation, instance area, and per-class size priors can improve performance in our depth estimation system. We choose this information to align with known depth cues in the biological domain: familiar size, relative size, and absolute size. 
    These cues relate the apparent size of an object to either the size of other, similarly sized objects, or to a known size prior of how large that object actually is in world coordinates. 
    Without this explicit information, the model must learn about these cues itself, and with any insufficiency in either model capacity or data availability, it will struggle to extract this information. Our method overcomes this issue, and provides a way to incorporate knowledge from outside the depth dataset.
    
    Our method is easily applicable to any depth estimation system, improves performance significantly over the baseline, and can be easily expanded to include information from further sources if required. We believe that introducing this information to other, more complex systems would still yield an increase in performance. Further, we believe that with the use of more depth-focused language data and with better models of external knowledge, significant improvements may be obtained. In our future work, these experiments will be undertaken. In this way we expect to be able to prove in the future the general applicability of this method, and therefore to provide further evidence that monocular depth estimation systems may be making use of cues identified in the biological domain.

\bibliographystyle{IEEEtran}
\bibliography{references}

\end{document}